\documentclass[letterpaper]{article} 
\usepackage{aaai25}  
\usepackage{times}  
\usepackage{helvet}  
\usepackage{courier}  
\usepackage[hyphens]{url}  
\usepackage{graphicx} 
\usepackage{natbib}  
\usepackage{caption} 
\usepackage{algorithm}
\usepackage{algorithmic}

\usepackage{bbding}
\usepackage{url}            
\usepackage{booktabs}       
\usepackage{amsfonts}       
\usepackage{nicefrac}       
\usepackage{xcolor}         
\usepackage{lipsum} 
\usepackage{amsmath}
\usepackage{amssymb} 
\usepackage{multirow}
\usepackage{relsize}
\usepackage{array}
\usepackage{graphicx}
\usepackage{colortbl}
\usepackage{pifont}
\usepackage{graphicx}
\usepackage{hhline}
\usepackage{boldline}
\usepackage{float}
\usepackage{subcaption}
\usepackage{enumitem}

\usepackage{newfloat}
\usepackage{listings}
\DeclareCaptionStyle{ruled}{labelfont=normalfont,labelsep=colon,strut=off} 
\floatstyle{ruled}
\newfloat{listing}{tb}{lst}{}
\floatname{listing}{Listing}
\title{\textit{SeFAR}: Semi-supervised Fine-grained Action Recognition with \\ Temporal Perturbation and Learning Stabilization}
\author{
    Yongle Huang\textsuperscript{\rm 1,2}\equalcontrib,
    Haodong Chen\textsuperscript{\rm 1,2}\equalcontrib,
    Zhenbang Xu\textsuperscript{\rm 1},
    Zihan Jia\textsuperscript{\rm 3},
    Haozhou Sun\textsuperscript{\rm 4},
    Dian Shao\textsuperscript{\rm 1}\thanks{Corresponding Author}\\
}
\affiliations{
    \textsuperscript{\rm 1}Unmanned System Research Institute, Northwestern Polytechnical University, Xi’an, China\\
    \textsuperscript{\rm 2}School of Automation, Northwestern Polytechnical University, Xi’an, China\\
    \textsuperscript{\rm 3}School of Computer Science, Northwestern Polytechnical University, Xi’an, China\\
    \textsuperscript{\rm 4}School of Software, Northwestern Polytechnical University, Xi’an, China\\

    \{yonglehuang, chd\}@mail.nwpu.edu.cn, shaodian@nwpu.edu.cn
}

\usepackage{bibentry}

\begin{document}

\maketitle

\begin{abstract}
Human action understanding is crucial for the advancement of multimodal systems. While recent developments, driven by powerful large language models (LLMs), aim to be \textit{general} enough to cover a wide range of categories, they often overlook the need for more \textit{specific} capabilities. In this work, we address the more challenging task of Fine-grained Action Recognition (FAR), which focuses on detailed semantic labels within shorter temporal duration (\textit{e.g.}, ``salto backward tucked with 1 turn").
Given the high costs of annotating fine-grained labels and the substantial data needed for fine-tuning LLMs, we propose to adopt semi-supervised learning (SSL). Our framework, \textbf{SeFAR}, incorporates several innovative designs to tackle these challenges. Specifically, to capture sufficient visual details, we construct \textit{Dual-level temporal elements} as more effective representations, based on which we design a new strong augmentation strategy for the Teacher-Student learning paradigm through involving \textit{moderate temporal perturbation}. 
Furthermore, to handle the high uncertainty within the teacher model's predictions for FAR, we propose the \textit{Adaptive Regulation} to stabilize the learning process.
Experiments show that SeFAR achieves state-of-the-art performance on two FAR datasets, FineGym and FineDiving, across various data scopes. It also outperforms other semi-supervised methods on two classical coarse-grained datasets, UCF101 and HMDB51. Further analysis and ablation studies validate the effectiveness of our designs.
Additionally, we show that the features extracted by our SeFAR could largely promote the ability of multimodal foundation models to understand fine-grained and domain-specific semantics. 
\textit{Code \& Datasets: \textcolor{magenta}{\url{https://github.com/KyleHuang9/SeFAR}}}.

\end{abstract}

\section{Introduction}

\begin{figure}[t]
    \centering
   \includegraphics[width=\linewidth]{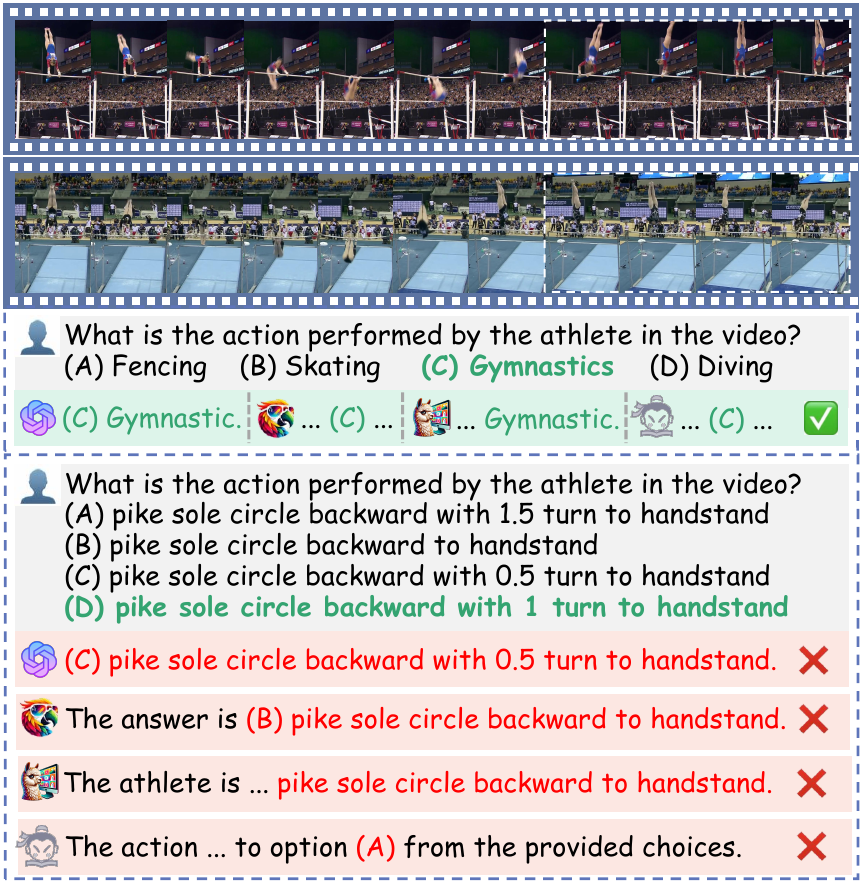}
   \vspace{-1.6em}
    \captionof{figure}{\textbf{Fine-grained Action Instances.} The two samples are drawn from the FineGym~\cite{shao2020finegym} dataset, specifically the \textit{``pike sole circle backward with 0.5 turn to handstand"} at the top and the \textit{``... 1 turn ..."} at the bottom. We further test popular MLLMs on the bottom instance for both coarse-grained and fine-grained: GPT-4V~\cite{openai2024gpt4vsystemcard}, VideoChat2~\cite{li2024mvbench}, VideoLLaVA~\cite{lin2023video}, and InternLM-XComposer-2.5~\cite{internlmxcomposer2_5}. 
    }
    \label{fig:finegrained}
    \vspace{-1.6em}
\end{figure}

Understanding videos has attracted increasing attention as videos contain vivid visual information and rich temporal dynamics absent in text and images.
In the past year, we have seen remarkable progress in multimodal large language models (MLLMs)~\cite{chen2023minigpt, li2024mvbench, li2023videochat, lin2023video}, aiming at acquiring more general and comprehensive abilities. 
However, as pointed out by recent studies~\cite{zhao2024videoprism,yuan2023videoglue}, chasing generality may sacrifice some task-specific performance, which motivates us to delve into a perpendicular direction:
focus on more \textit{specific} tasks to promote the fine-grained understanding ability of models.

Specifically, we focus on Fine-grained Action Recognition (FAR), a challenging human-centric video understanding task.
To explain, classical action recognition~\cite{xiong2021multiview, xiao2022learning, dave2023timebalance, xing2023svformer} only demands the model to provide relatively coarse-grained category such as \textit{``gymnastics"},
while FAR aims to provide more detailed, specific, and semantically accurate descriptions as \textit{``pike sole circle backward with 0.5 turn to handstand"}.
To demonstrate the difficulty of this task, we evaluate four powerful MLLMs~\cite{openai2024gpt4vsystemcard, li2024mvbench, lin2023video, internlmxcomposer2_5}, as shown in Fig.~\ref{fig:finegrained}. 
Unfortunately, they all fail to correctly recognize the fine-grained semantics of the given action.
In such a sense, FAR holds significance in further enhancing the capability of MLLM 
~\cite{driess2023palm, vemprala2024chatgpt}, especially in application scenes requiring more accurate and professional information.


However, limited research on FAR not only owes to its higher demands for method design but also the dataset construction~\cite{shao2020finegym, xu2022finediving}.
For example, providing annotations such as \textit{``5237D with 3.5 twists"}~\cite{xu2022finediving} requires adequate expert knowledge, huge annotation time, and large checking efforts to ensure the quality~\cite{shao2020finegym}. 
This leads to the scarcity of fine-grained labels and makes it difficult to directly re-train or fine-tune large models with huge annotated data.
Keep this in mind,
we further adopt the semi-supervised learning (SSL) setting, where only a small percentage of labeled data is needed~\cite{zhu2005semi}.
Consequently, targeting semi-supervised FAR, besides those intrinsic challenges from both sides, we have to tackle intractable \textit{new challenges} that emerged when combined.
Specifically,
FAR needs enough visual details, effective information aggregation, and a comprehensive understanding of temporal dynamics~\cite{shao2020finegym, xu2022finediving, li2022dynamic, tang2023m3net}.
For SSL, the core is to equip the unlabeled data with stable and reasonable supervision (\textit{e.g.}, pseudo-labels)~\cite{sohn2020fixmatch, zhu2005semi, kurakin2020remixmatch}.
However, when training a semi-supervised FAR model,
the generated pseudo-labels may not be reliable, since FAR is rather challenging, making the whole learning process easily collapse.

In this paper, we propose a novel framework, \textbf{SeFAR}, to address the above challenges.
Due to the semi-supervised setting, SeFAR is developed based on the FixMatch~\cite{sohn2020fixmatch} SSL paradigm, including the weak-to-strong consistency regularization and the Teacher-Student setup, as shown in Fig.~\ref{fig:pipeline}.
Moreover, there are also delicately designed strategies and modules incorporated in SeFAR:
\ding{182} First, to effectively mine adequate and useful data for FAR, a \textit{dual-level information modeling} strategy is proposed. This process combines both fine-grained temporal elements with the temporal context to effectively capture multi-granular temporal information, enhancing the ability to discriminate subtle actions in the video.
\ding{183} Then, to construct weak-strong contrast data pairs more tailored for FAR which differs from the traditional spatial-only augmentations~\cite{yun2019cutmix, devries2017improved, kurakin2020remixmatch}, we highlight the significance of temporal dynamics and design a new strong augmentation strategy. Specifically, we introduce \textit{moderate temporal perturbation} into the fine-grained temporal elements achieved previously, while keeping the temporal order of context element. 
\ding{184} Moreover, in order to provide reliable pseudo-labels for unlabeled data even when the Teacher model suffers from unstable predictions,
we design an \textit{Adaptive Regulation} to stabilize the training process by calculating coefficients to adjust the losses.
In addition, 
to directly tackle the problems outlined in Fig.~\ref{fig:finegrained}, we adhere to the standard MLLM framework, which includes a vision encoder, a language encoder, and an alignment adapter. By incorporating our SeFAR model as an innovative video encoder, we observe that all MLLMs perform better on FAR, as shown in Tab.~\ref{tab:mllm}.


To summarize, our contributions are as follows:
\begin{itemize}[leftmargin=*]
    \item To the best of our knowledge, this is the first work to explore the highly challenging task of \textbf{Se}mi-supervised \textbf{F}ine-grained \textbf{A}ction \textbf{R}ecognition and an effective framework \textbf{SeFAR} is proposed for this purpose, which is based on the FixMatch paradigm but incorporates a new augmentation strategy to form the weak-to-strong data pairs;
    \item Moreover, SeFAR incorporates several innovative designs to address specific challenges, including the dual-level temporal elements modeling, careful involvement of moderate temporal perturbation, as well as the adaptive regulation for a steady learning process;
    \item SeFAR achieves state-of-the-art performance on both fine-grained (FineGym, FineDiving) and coarse-grained action recognition datasets (UCF101, HMDB51), demonstrating its effectiveness. Additional analysis shows that SeFAR could also serve as a powerful visual encoder to assist current MLLMs in domain-specific scenes.
\end{itemize}



\section{Related Work}
\subsubsection{Fine-grained Action Recognition (FAR).}
FAR aims to differentiate between similar human actions at a finer semantic granularity (\textit{e.g.},\textit{``switch leap with 0.5 turns" vs. ``split jump with 1 turn"}), while coarse-grained actions~\cite{zhou2018temporal, carreira2017quo, xu2022cross, yang2020temporal, wang2018temporal}, stop at the granularity of \textit{``gymnastics"}. To achieve this, abundant and subtle motion details are extremely desired~\cite{shao2020finegym}. 
There are several pioneer works~\cite{li2022dynamic, leong2022combined, leong2021joint, tang2023m3net, hong2021video, wang2021few} to tackle the problem of FAR. However, they have predominantly focused on fully supervised or few-shot learning. Among them, LCDC~\cite{mac2019learning} capture local spatio-temporal features, HAAN~\cite{li2022weakly} use hierarchical modeling with atomic actions and visual concepts, while M$^3$Net~\cite{tang2023m3net} implement multi-view encoding, matching, and fusion. 
Distinct from the above works,
we propose to address a more challenging task and propose the first semi-supervised FAR framework, SeFAR, integrating with the \textit{dual-level temporal elements modeling}, which tackles the subtle inter-class differences but also contends with limited annotations.
\vspace{-0.2em}


\begin{figure*}[!t]
    \vspace{-0.6em} 
    \centering
    \includegraphics[width=1\textwidth]{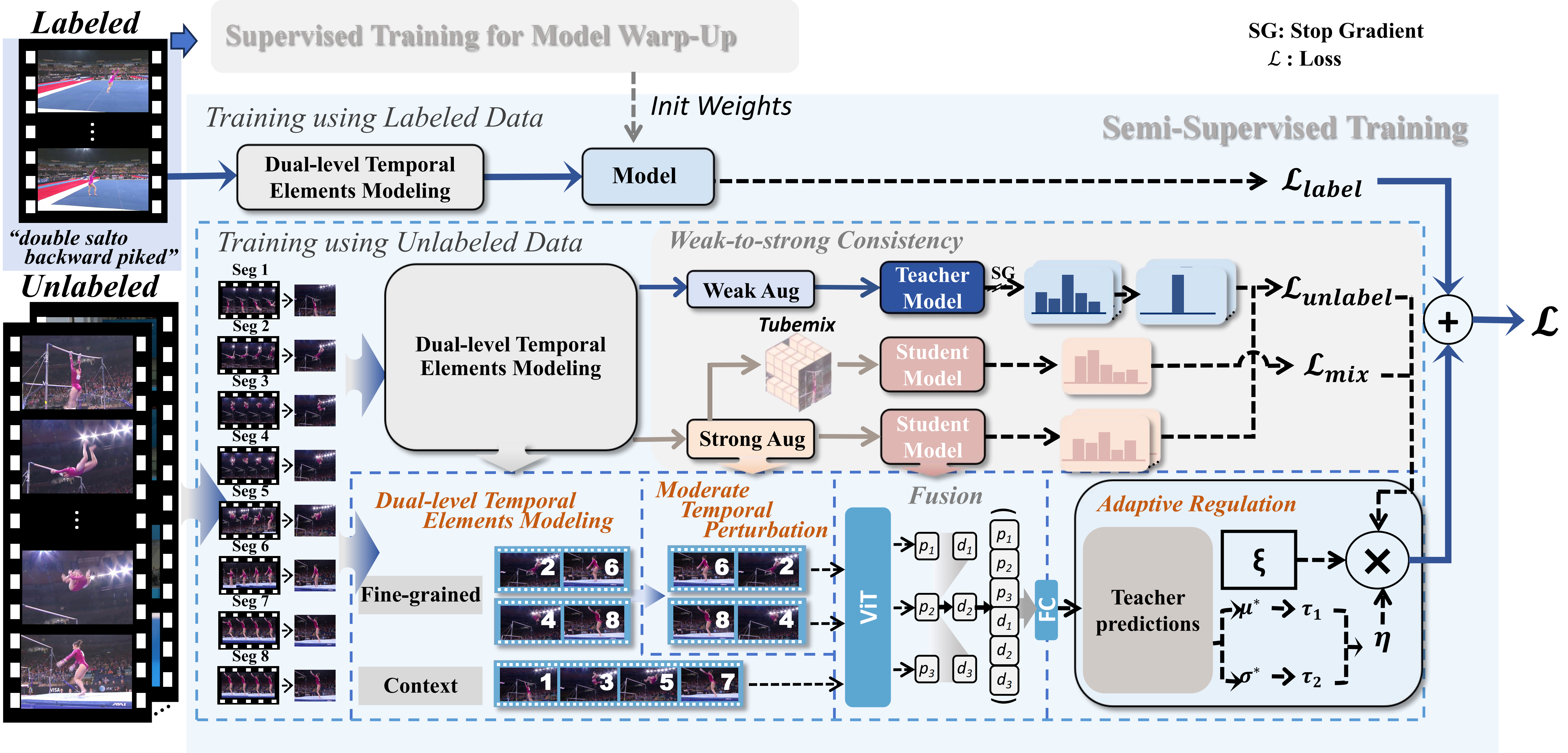}
    \vspace{-2em}
    \captionof{figure}{\label{fig:pipeline}
    \textbf{Overview of SeFAR pipeline.} 
    We target Semi-supervised FAR, assuming most input samples are unlabeled. 
    During unsupervised learning, SeFAR adopts \textit{dual-level temporal elements modeling} and performs augmentation in two manners (`Weak' \textit{vs.} `Strong'). Strongly augmented/distorted samples by \textit{moderate temporal perturbation} are used by the student model, while the teacher model offers pseudo-labels based on weakly augmented samples. Consistency is enforced through loss minimization ($\mathcal{L}_{un}$). The unsupervised loss is further adjusted by our proposed \textit{Adaptive Regulation}. The framework is trained with a weighted combination of supervised $\mathcal{L}_{sup}$ and unsupervised $\mathcal{L}_{un}$ losses.
    }
   \vspace{-1em}
\end{figure*}

\vspace{-0.2em}
\subsubsection{Data Augmentation in Semi-supervised Learning (SSL).}
Data augmentation plays an essential role in SSL, serving as one of the two core components of the FixMatch~\cite{sohn2020fixmatch}-based paradigm, specifically \textit{consistency regularization} achieved through both strong and weak data augmentation. This has been previously demonstrated. For instance, \cite{xie2020unsupervised} emphasizes that a robust model should withstand variations in input examples or hidden states. However, most existing semi-supervised video action recognition studies~\cite{xu2022cross, xiong2021multiview, xiao2022learning, dave2023timebalance} focus primarily on spatial augmentations achieved through image-based strategies (\textit{e.g.}, Cutmix~\cite{yun2019cutmix}, Cutout~\cite{devries2017improved}, or their variants~\cite{kurakin2020remixmatch, cubuk2020randaugment}). We argue that temporal augmentation is equally important inspired by~\cite{xing2023svformer}, especially in FAR, as spatial augmentations can often disrupt critical information within actions. To address this, we design a new temporal augmentation strategy, \textit{moderate temporal perturbation}. Furthermore, to maintain stability in the \textit{pseudo-labeling} process, another core component of the FixMatch-based paradigm, we have developed the \textit{Adaptive Regulation} during training.

\section{Methodology}

 To tackle the challenging task of semi-supervised fine-grained action recognition,
 we propose the SeFAR framework, and the complete pipeline is shown in Fig.~\ref{fig:pipeline}.
Before delving into specific details, we first elaborate on the preliminaries about semi-supervised learning, especially the FixMatch~\cite{sohn2020fixmatch} paradigm. 
\vspace{-0.3em}
\subsection{Preliminaries}
\label{sec:pre}

\subsubsection{\ding{113} Teacher \textit{vs.} Student Model.}
A line of SSL frameworks adopts the Teacher-Student setting, where the \textit{Teacher} provides pseudo-labels to supervise the \textit{Student} model.
Instead of directly sharing weights between teacher and student models~\cite{sohn2020fixmatch}, we adopt an average of consecutive student models to obtain a ``Mean teacher", whose effectiveness has been verified~\cite{tarvainen2017mean}. 
Formally, at a given time step, the weights of the \textit{Teacher} model, $\theta_{t}$, is updated as an exponential moving average of the student weights $\theta_{s}$:
{\setlength\abovedisplayskip{3pt}
\setlength\belowdisplayskip{3pt}
\begin{equation} 
\theta_{t}  \longleftarrow \omega \theta_{s} + (1-\omega)\theta_{t}. \label{4} 
\end{equation}}

\noindent As pointed out in~\cite{xing2023svformer}, such EMA-Teacher is more suitable and stable for human action recognition.

\vspace{-0.2em}
\subsubsection{\ding{113} Weak \textit{vs.} Strong Augmentation.}
One core component within FixMatch~\cite{sohn2020fixmatch} is the construction of contrastive data pairs to facilitate consistency regularization.
This involves the incorporation of both strong and weak augmentations, wherein the term \textit{``augmentation"} here means \textit{``distortion"} rather than ``enhancement", contrary to intuition.
Specifically, strong augmentation ($\mathcal{A}_{strong}$) usually causes significant perturbation to the original data and thus serves as the input for the Student model,
while the $\mathcal{A}_{weak}$ produces moderately distorted data samples for the Teacher model to derive better predictions, as demonstrated in the center part of Fig.~\ref{fig:pipeline}.

%

\vspace{-0.2em}
\subsubsection{\ding{113} Learning by Labeled \textit{vs.} Unlabeled Data.}
In the SSL setting, only a small portion of data is annotated, denoted by $\{x_{i},y_{i}\}_{i=1}^{\mathcal{B}_l}$.
The left ${\mathcal{B}_{u}}$ samples, $\{x_{j}\}_{j=1}^{\mathcal{B}_{u}}$, are all unlabeled.
Usually the labeling ratio $\alpha =\frac{ \mathcal{B}_l }{\mathcal{B}_l + \mathcal{B}_u}  $  is small (\textit{e.g.}, $0.1$).
Learning based on the labeled data is straightforward by minimizing the 
cross-entropy loss between model predictions $Pred(x_i)$ and labels $y_i$:
{\setlength\abovedisplayskip{3pt}
\setlength\belowdisplayskip{3pt}
\begin{equation}
\mathcal{L}_{sup} = \frac{1}{\mathcal{B}_l}\sum^{\mathcal{B}_l}_{i=1}\mathcal{H}(y_i, Pred(x_i)). \label{eq2} 
\end{equation}}

\noindent However, for the unlabeled data $x_j$, there is no supervision.
To solve this, we generate pseudo-labels from the Teacher model predictions $\mathcal{F}^{T}$, and then 
calculate the unsupervised loss as follows:
{\setlength\abovedisplayskip{4pt}
\setlength\belowdisplayskip{3pt}
\begin{equation}
  \begin{aligned}
\hat{y_j} &= max( {\mathcal{F}}_{t} (\mathcal{A}_{weak}(x_j)),\\
\mathcal{L}_{un} &= \frac{1}{\mathcal{B}_{u}}\sum^{\mathcal{B}_u}_{j=1}\mathbf{1}( \hat{y_j} > \tau)\mathcal{H}(\hat{y_j}, \mathcal{F}_s(\mathcal{A}_{Strong}(x_j))), \label{eq3} 
\end{aligned}  
\end{equation}}

\noindent where $\tau$ is the predefined threshold for confidence scores and $\mathbf{1}$ denotes the indicator function.
The whole pipeline is trained using both losses, weighted by hyperparameters, 
{\setlength\abovedisplayskip{3pt}
\setlength\belowdisplayskip{3pt}
\begin{equation} 
\mathcal{L}= \gamma_{1} \mathcal{L}_{sup} + \gamma_2 \mathcal{L}_{un}. \label{eq4} 
\end{equation}}

\subsection{The SeFAR Framework}
\label{sec:framework}
In this work, we focus on the task of Fine-grained Action Recognition (FAR) in the Semi-Supervised Learning (SSL) setting.
This new task brings unprecedented challenges, including: 
\ding{182} How to mine abundant and detailed visual information for differentiating subtle differences between fine-grained actions?
\ding{183} How to adapt the original SSL strategies, \textit{e.g.}, consistency regularization, to fit the ``temporal-matters" FAR task?
\ding{184} How to deal with the unstable pseudo-labels since the model hesitates between appearance-similar action samples?
In the following paragraphs, we will introduce specific designs to address the above challenges.

\vspace{-0.2em}
\subsubsection{\ding{109} Dual-level Temporal Elements.}
Given a fine-grained action video with $N$ frames,  we first trim it into $K$ segments~\cite{wang2016temporal}, and randomly sample one frame in each segment, obtaining a frame sequence $\{f_1, ..., f_K\}$ to represent the video.
Since in FAR, the high similarity is shared in large parts of visual content (\textit{e.g.}, scenes, objects), models are usually required to perceive subtle changes and abundant details for accurate discrimination.
To achieve this,  we propose to construct several small temporal elements $p_i$, where \textit{``small"}  means the size $L$ (\textit{i.e.}, the number of containing frames) of $p_i$ is moderate. 
Intuitively, a small value of $L$ could help the model focus on quick and subtle changes, since details are usually missed when going through too many frames.
Given K frames, the sampling step is  $\lfloor \frac{K}{L} \rfloor$.
After sampling $M$ times, we could get a set of temporal elements with the same temporal lengths, denoted by:
{\setlength\abovedisplayskip{3pt}
\setlength\belowdisplayskip{3pt}
\begin{equation}
  \{ p_i \}_{i=1}^M, \quad |p_i| = L.  
\end{equation}}

\noindent Besides these temporally fine-grained elements, we also propose to obtain a context element $p^{\textbf{context}}$ to encode long-term information and macro temporal dynamics.
$p^{\textbf{context}}$ is composed of more frames, usually two times more than the fine-grained temporal elements $p_i$.
Such dual-level information modeling ensures that multi-granular information is preserved.
As a result, we obtain an effective representation of the input video, denoted by $\{p_1, ...p_M, p^{\textbf{context}} \}$.

\vspace{-0.2em}
\subsubsection{\ding{109} Perturbation of Fine-grained Temporal Elements.}
Adopting the FixMatch~\cite{sohn2020fixmatch} based semi-supervised learning setting, one key problem is \textit{``how to form the weak-to-strong augmentation pair for consistency regularization".}
For weak augmentation, we could use random horizontal flipping or random scaling, since it largely preserves both spatial and temporal original information.
Unfortunately, as pointed out in~\cite{xing2023svformer}, strong augmentation designed for images is insufficient for video tasks, since it fully ignores the temporal dynamics evolving in videos.
For the challenging FAR task, temporal variations are even more crucial and require the extreme attention of the model.
Therefore, to design more effective strong augmentation strategy $\mathcal{A}_{strong}$ for FAR, we emphasize the following core insights:
\ding{172} the proposed $\mathcal{A}_{strong}$ should make perturbations to the most crucial part of the data that we want the model to attend to~\cite{sohn2020fixmatch, xie2020unsupervised, kurakin2020remixmatch};
\ding{173} Employing $\mathcal{A}_{strong}$ should not affect the semantic distinctiveness of action categories.

Therefore, combing with the above dual-level temporal modeling strategy, we propose a new strong augmentation operation through introducing temporal perturbation $\psi$ into the fine-grained temporal elements $\{ p_i \}$.
We experiment with different implementations of $\psi$, and the final choice is simple but effective: \textit{reversing the frame order}. Specifically, we have:
{\setlength\abovedisplayskip{3pt}
\setlength\belowdisplayskip{3pt}
\begin{align}
    \mathcal{A}_{strong}(\, \{ p_i \}_{i=1}^{M} \,) = \{\; \overleftarrow{p_i} \; \}_{i=1}^{M},  \quad \overleftarrow{p_i} = \psi (p_i)
\end{align}}

\noindent Note that for the temporal context element $p^{\textbf{context}}$, the temporal order is preserved, which ensures the temporal directionality to be inherent in actions (\textit{e.g.}, \textit{``giant circle backward" vs. ``giant circle forward"}, etc.),
as shown in the bottom-left of Fig.~\ref{fig:pipeline}.
Our augmentation strategy introduces moderate temporal perturbation compared with total shuffling,
and it also outperforms previous strategies, \textit{e.g.}, temporal warping~\cite{xing2023svformer}, as shown in Tab.~\ref{tab:sub3.2}.

\begin{figure}
    \centering
    \vspace{-0.4em} 
   \includegraphics[width=\linewidth]{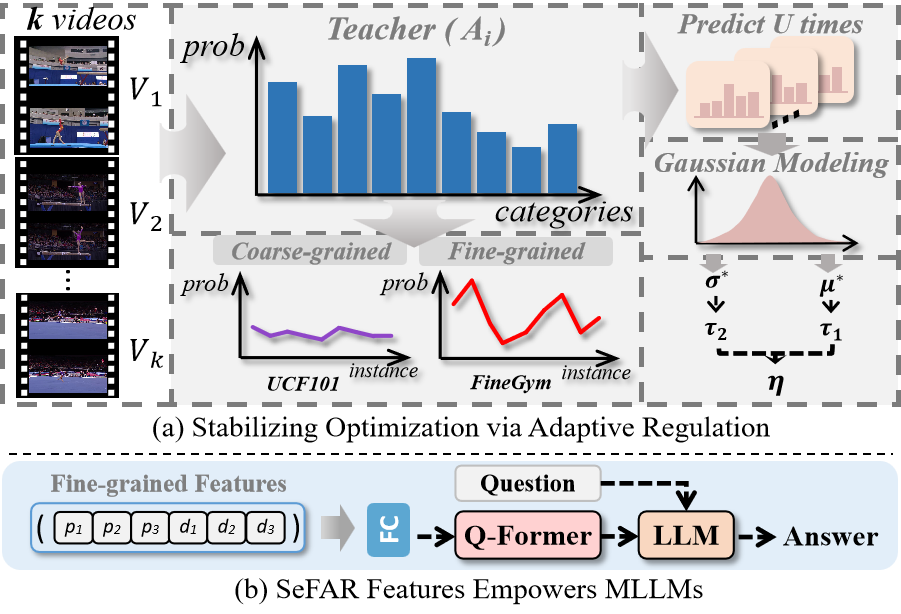}
   \vspace{-2em}
    \captionof{figure}{\label{fig:prediction}
    \textbf{(a)} For $K$ unlabeled videos, the Teacher model predicts each video multiple times to capture the distribution of predictions, which shows less variability on coarse-grained data and more on fine-grained data. An adaptive coefficient $\eta$ is calculated from the mean and variance of the distribution to stabilize training.
    \textbf{(b)} MLLM construction pipeline with SeFAR's fine-grained features.
    }
    \vspace{-1.2em}
\end{figure}

\begingroup
\begin{table*}[ht]
\vspace{-0.6em}
  \caption{\textbf{Comparison with state-of-the-art semi-supervised action recognition methods on fine-grained datasets.} We employ SeFAR with a sampling combination of \{2-2-4\}. The primary evaluation metric is top-1 accuracy. In this table, ``V" within ``Input" denotes RGB video, while ``G" represents temporal gradients. ``ImgNet" indicates the utilization of models pre-trained on ImageNet~\cite{russakovsky2015imagenet}, while ``\#F" signifies the number of input frames. The labeling rates of the data are indicated by ``5\%", ``10\%", and ``20\%" in the datasets. The best results are highlighted in \textbf{Bold}, and the second-best \underline{Underlined}.}
  \vspace{-0.6em}
  \setlength{\tabcolsep}{1.5pt} 
  \begin{subtable}{\textwidth}
  \renewcommand{\arraystretch}{1}
  \centering
  \small
   \begin{tabular}{lcccccc|cccccc}
     \hlineB{2.5}
     \multirow{2}{*}{\textbf{Method}} & \multirow{2}{*}{\textbf{Backbone}} & \multirow{2}{*}{\textbf{Input}} & \multirow{2}{*}{\textbf{ImgNet}} & \multirow{2}{*}{\textbf{Params}} & \multirow{2}{*}{\textbf{\#F}} & \multirow{2}{*}{\textbf{Epoch}} & \multicolumn{2}{c}{\textbf{Gym99}} & \multicolumn{2}{c}{\textbf{Gym288}} &\multicolumn{2}{c}{\textbf{Diving}}\\
     \cline{8-13}
     & & & & & & & \textbf{5\%} & \textbf{10\%} & \textbf{5\%} & \textbf{10\%} & \textbf{5\%} & \textbf{10\%}\\
     \hlineB{2}
     MemDPC (ECCV'20)~\cite{han2020memory} &  3D-ResNet-18 & V & \ding{55} & 15.4M & 16 & 500 & 10.8 & 24.1 & 14.5 & 21.3 & 54.3 & 62.0\\
     LTG (CVPR'22)~\cite{xiao2022learning} & 3D-ResNet-18 & VG & \ding{55} & 68.3M & 8 & 180 & 34.3 & 45.8 & 16.2 & 38.7 & 59.8 & 64.3\\
     SVFormer (CVPR'23)~\cite{xing2023svformer} & ViT-B & V & \ding{51} & 121.4M & 8 & 30 & 31.4 & 47.9 & 21.3 & 39.6 & 59.1 & 70.8\\
     \hline
     \rowcolor{cyan!10}
     SeFAR-S (Ours) & VIT-S & V & \ding{51} & 31.2M & 8 & 30 & \underline{36.7}  & \underline{56.3} & \underline{27.8} & \underline{46.9} & \underline{72.2} & \underline{78.4} \\
     \rowcolor{cyan!10}
     SeFAR-B (Ours) & VIT-B & V & \ding{51} & 122.1M & 8 & 30 & \textbf{39.0}  & \textbf{56.9} & \textbf{28.3} & \textbf{48.1} & \textbf{72.8} & \textbf{80.9} \\
     \hlineB{2.5}
   \end{tabular}
   \caption{Results of elements across all events.}
   \label{tab:sub1.1}
   \end{subtable}
   \hfill
   \setlength{\tabcolsep}{5.9pt}
    \begin{subtable}{0.5\linewidth}
    \renewcommand{\arraystretch}{1}
    \centering
    \small
    \begin{tabular}{lcccccc}
     \hlineB{2.5}
     \multirow{2}{*}{\textbf{Method}} & \multicolumn{2}{c}{\textbf{UB}} & \multicolumn{2}{c}{\textbf{FX}} & \multicolumn{2}{c}{\textbf{10m}}\\
     \cline{2-7}
     & \textbf{10\%} & \textbf{20\%} & \textbf{10\%} & \textbf{20\%} & \textbf{10\%} & \textbf{20\%}\\
     \hlineB{2}
     MemDPC & 20.7 & 19.1 & 13.8  & 15.9 & 65.4 & 71.2\\
     LTG & 50.5 & 60.5 & 19.6 & 21.6 & 75.2 & 83.5\\
     SVFormer & 52.9 & 66.8 & 20.1 & 28.8 & 73.8 & 85.9\\
     \hline
     \rowcolor{cyan!10}
     SeFAR-S (Ours) & \underline{56.9} & \underline{73.8}& \underline{23.8} & \underline{42.9} & \underline{85.5} & \underline{94.0}\\
     \rowcolor{cyan!10}
     SeFAR-B (Ours) & \textbf{58.5} & \textbf{75.5}& \textbf{27.6} & \textbf{44.2} & \textbf{87.4} & \textbf{94.6}\\
     \hlineB{2.5}
   \end{tabular}
   \caption{Results of elements within an event.}
   \label{tab:sub1.2}
   \end{subtable}
   \hfill
   \setlength{\tabcolsep}{5.8pt}
    \begin{subtable}{0.5\linewidth}
    \renewcommand{\arraystretch}{1}
    \small
    \begin{tabular}{lcccccc}
     \hlineB{2.5}
     \multirow{2}{*}{\textbf{Method}} & \multicolumn{2}{c}{\textbf{UB-S1}} & \multicolumn{2}{c}{\textbf{FX-S1}} & \multicolumn{2}{c}{\textbf{5253B}}\\
     \cline{2-7}
     & \textbf{10\%} & \textbf{20\%} & \textbf{10\%} & \textbf{20\%} & \textbf{10\%} & \textbf{20\%}\\
     \hlineB{2}
     MemDPC & 17.2 & 21.1 & 15.4 & 20.1 & 82.2 & 89.5\\
     LTG & 21.3 & 29.7 & 14.6 & 19.3 & 64.6 & 76.9\\
     SVFormer & 28.9 & 47.3 & 18.8 & 22.5 & 86.6 & 90.1\\
     \hline
     \rowcolor{cyan!10}
     SeFAR-S (Ours) & \underline{36.6} & \underline{55.3} & \underline{19.2} & \underline{25.5} & \underline{96.4} & \underline{97.3} \\
     \rowcolor{cyan!10}
     SeFAR-B (Ours) & \textbf{37.1} & \textbf{56.8} & \textbf{20.1} & \textbf{26.5} & \textbf{97.0} & \textbf{97.8} \\
     \hlineB{2.5}
   \end{tabular}
   \caption{Results of elements within a set.}
   \label{tab:sub1.3}
   \end{subtable}
   \label{tab:1}
   \vspace{-1.6em}
\end{table*}
\endgroup

\vspace{-0.2em}
\subsubsection{\ding{109} Stabilizing Optimization via Adaptive Regulation.}
As mentioned, due to the challenging intrinsic of FAR,
models usually swayed precariously between categories with subtle differences.
During experiments, the greater the uncertainty of the model's predictions, the less reliable the model's predictions are.
Such unstable predictions of the teacher model will result in ambivalent and invalid pseudo-labels for the student,
making the whole learning process suffer.
To solve this,
we first let the \textit{Teacher} model generate predictions $U$ times ($U$ is set to 10 in experiments) for a given unlabeled video, and these predictions may vary largely.
Then, based on these, we calculate the mean prediction confidence and standard deviation for each category.
For the $i^{th}$ prediction, the predicted probability across all categories constitutes a probability distribution. From this distribution, we can obtain the maximum prediction confidence value $\mu^{i}$ and calculate its standard deviation $\sigma^{i}$.
We select the highest confidence value $\mu^* = max(\mu^{i})$,
along with its corresponding standard deviation $\sigma^*$ (see Fig.~\ref{fig:prediction}).

Based on such $\mu^*$ and $\sigma^*$,
we propose to calculate the dynamic coefficients $\tau_1$ and $\tau_2$ to obtain $\eta$, which is further used for adjusting losses derived from unlabeled samples:
{\setlength\abovedisplayskip{3pt}
\setlength\belowdisplayskip{3pt}
\begin{equation}
  \begin{aligned}
\tau_1 & =  sigmoid(e^{\mu^*} - e), \\
\tau_2 & = (sigmoid(\frac{1}{\beta\sigma^*+\epsilon})-0.5),
\end{aligned}  
\end{equation}}

\noindent where $\beta$ is related to the model dropout and $\epsilon$ is a steady parameter.
To elaborate,
$\tau_1$ will increase rapidly as $\mu^*$ increases, which enhances high-confidence predictions,
while on the other hand, $\tau_2$ suppresses the unstable predictions (\textit{i.e.}, with high standard deviation $\sigma$).
The obtained adaptive coefficient $\eta = \tau_1 \cdot \tau_2,$ is more flexible and beneficial than a predefined hyperparameter. 
Additionally, for unlabeled data, we also adopt the mixing strategy as in SVFormer~\cite{xing2023svformer}, where the mixture of two unlabeled samples, $\lambda x_1  + (1-\lambda)x_2$, could also serve as input, and the supervision is correspondingly obtained as a mixed version (Details could be found in~\cite{xing2023svformer}).
Here for adjusting $\mathcal{L}_{mix}$, we achieve its coefficient in a similar mixed manner, denoted by $\eta^{'}=\lambda\eta_{1}+(1-\lambda)\eta_{2}$, where $\eta_{1}, \eta_{2}$ are individually calculated based on $x_1$ and $x_2$. 
Finally, the total loss of the whole SeFAR framework is as follows:
{\setlength\abovedisplayskip{3pt}
\setlength\belowdisplayskip{3pt}
\begin{equation}
\mathcal{L} = \mathcal{L}_{sup} + \xi(\eta\mathcal{L}_{un} + 
 \eta^{'} \mathcal{L}_{mix}),
 \label{eq-loss} 
\end{equation}}

\noindent where $\xi = sin(\frac{n}{M_n})$ is a warmup coefficient calculated using the current epoch number $n$ and the max epoch $M_n$.

\vspace{-0.2em}
\subsubsection{\ding{109} SeFAR Empowers MLLMs.}
Efforts towards foundation models have led to the development of MLLMs, with vision being the primary modality~\cite{gao2024sphinx}. 
Although shown impressive general capabilities, they may fail in specific and more challenging tasks such as FAR, as illustrated in Fig.~\ref{fig:finegrained}.
This may largely be due to the systematic shortcomings in the visual part as analyzed in \cite{tong2024eyes}.
Given that our SeFAR is designed to be effective for FAR in semi-supervised scenarios, the question: \textit{``Could SeFAR benefit current MLLMs through providing better visual perception?"}
The answer is yes as supported by the results in Tab.~\ref{tab:mllm}. 
To elaborate,
in line with the typical MLLM framework, a frozen visual encoder is usually combined with a LLM. This setup facilitates multimodal functionality by aligning visual and textual features using an adaptor, \textit{e.g.}, Q-Former~\cite{li2023blip}. 
Given such a setting, we could use the features extracted by SeFAR to replace those provided by the original visual encoder as shown at the bottom of Fig.~\ref{fig:prediction}.
Similarly, by aligning the visual features with the textual domain and concatenating with text embeddings, we could feed them into the LLM to produce the answers. Results show that SeFAR features could lead to much better results compared to those used in original MLLM settings.

\begingroup
\setlength{\tabcolsep}{3.8pt}
\begin{table*}[t]
\vspace{-0.6em}
\renewcommand{\arraystretch}{1}
  \centering
  \caption{\textbf{Comparison with state-of-the-art semi-supervised action recognition methods on coarse-grained datasets.}``V" within ``Input" signifies RGB video, ``F" indicates optical flow, while ``G" denotes temporal gradients.}
  \centering
  \vspace{-0.8em}
  \small
   \begin{tabular}{lccccc|ccccc}
     \hlineB{2.5}
     \multirow{2}{*}{\textbf{Method}} & \multirow{2}{*}{\textbf{Backbone}} & \multirow{2}{*}{\textbf{Input}} & \multirow{2}{*}{\textbf{ImgNet}} & \multirow{2}{*}{\textbf{\#F}} & \multirow{2}{*}{\textbf{Epoch}} & \multicolumn{3}{c}{\textbf{UCF-101}}&\multicolumn{2}{c}{\textbf{HMDB-51}}\\
     \cline{7-11}
     & & & & & & \textbf{1\%} & \textbf{5\%} & \textbf{10\%} & \textbf{40\%} & \textbf{50\%} \\
     \hlineB{2}
     MT+SD (WACV'21)~\cite{jing2021videossl} & 3D-ResNet-18 & V & \ding{55} & 16 & 500 & - & 31.2 & 40.7 & 32.6 & 35.1  \\
     MvPL (ICCV'21)~\cite{xiong2021multiview} & 3D-ResNet-50 & VFG & \ding{55} & 8 & 600 & 22.8 & 41.2  & 80.5 & 30.5 & 33.9  \\
     TCLR (CVIU'22)~\cite{Dave_2022} & 3D-ResNet-18 & V & \ding{55} & 16 & 1200 & 26.9 & - & 66.1 & - & -  \\
     CMPL (CVPR'22)~\cite{xu2022cross} & R50+R50-1/4 & V & \ding{55} & 8 & 200 & 25.1 & - & 79.1 & - & - \\
     LTG (CVPR'22)~\cite{xiao2022learning} & 3D-ResNet-18 & VG & \ding{55} & 8 & 180 & - & 44.8  & 62.4 & 46.5 & 48.4  \\
     TimeBalance (CVPR'23)~\cite{dave2023timebalance} & 3D-ResNet-50 & V & \ding{55} & 8 & 250 & 30.1  & 53.3 & 81.1 & 52.6 & 53.9 \\
     \hline
     \rowcolor{yellow!10}
     SeFAR (Ours) & VIT-S & V & \ding{55} & 8  & 30 & 35.2 & 64.1 & 78.3 & 55.9 & 59.2 \\
     \hlineB{1.6}
     FixMatch (NeurlPS'20)~\cite{sohn2020fixmatch} & SlowFast-R50 & V & \ding{51} & 8 & 200 & 16.1 & - & 55.1 & - & -  \\
     MemDPC (ECCV'20)~\cite{han2020memory} & 3D-ResNet-18 & V & \ding{51} & 16 & 500 & - & - & 44.2 & - & -  \\
     ActorCM (CVIU'21)~\cite{zou2023learning} & R(2+1)D-34 & V & \ding{51} & 8 & 360 & - & 45.1 & 53.0 & 35.7 & 39.5  \\
     VideoSSL (WACV'21)~\cite{jing2021videossl} & 3D-ResNet-18 & V & \ding{51} & 16 & 500 & - & 32.4 & 42.0 & 32.7 & 36.2  \\
     TACL (TSVT'22)~\cite{9904603} & 3D-ResNet-50 & V & \ding{51} & 16 & 200 & -  & 35.6 & 55.6 & 38.7 & 40.2  \\
     L2A (ECCV'22)~\cite{gowda2022learn2augment} & 3D-ResNet-18 & V & \ding{51} & 8 & 400 & -  & -  & 60.1 & 42.1 & 46.3  \\
     SVFormer-S (CVPR'23)~\cite{xing2023svformer} & ViT-S & V & \ding{51} & 8 & 30 & 31.4 & -  & 79.1 & 56.2 & 58.2  \\
     SVFormer-B (CVPR'23)~\cite{xing2023svformer} & ViT-B & V & \ding{51} & 8 & 30 & 46.1 & -  & 84.6 & 59.9 & 64.3  \\
     \hline
     \rowcolor{cyan!10}
     SeFAR (Ours) & VIT-S & V & \ding{51} & 8  & 30 & 46.0 & 73.2 & 84.3 & 58.5 & 62.9 \\
     \rowcolor{cyan!10}
     SeFAR (Ours) & VIT-B & V & \ding{51} & 8  & 30 & \textbf{50.3} & \textbf{77.6} & \textbf{87.0} & \textbf{61.5} & \textbf{65.7} \\
     \hlineB{2.5}
   \end{tabular}
  \label{tab:2}
  \vspace{-0.3cm}
\end{table*} 
\endgroup

\begin{figure*}[t]
    \vspace{-0.4em} 
    \centering
    \includegraphics[width=\linewidth]{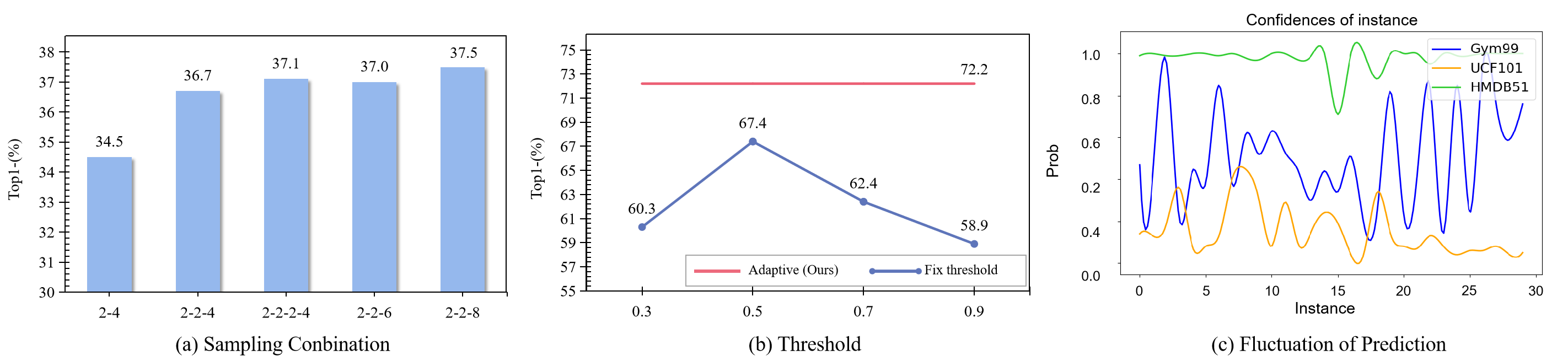}
    \vspace{-1.9em}
    \captionof{figure}{\label{fig:threshold}
    \textbf{Ablation Studies.} We compare SeFAR-B with different sampling combinations on Gym-99 5\%, as illustrated on the \textbf{left}. We also contrast fixed threshold methods with our Adaptive Regulation strategy on FineDiving 5\% in the \textbf{middle}. On the \textbf{right} side, we demonstrate the fluctuation of predictions made by the Teacher model across different datasets.
    }
    \vspace{-1.2em}
\end{figure*}

\section{Experiment}
\subsection{Experiment Setup}
\label{sec:setup}
\subsubsection{Datasets and Evaluation.}
We perform evaluations on fine-grained datasets Gym99, Gym288~\cite{shao2020finegym}, and FineDiving~\cite{xu2022finediving}, as well as coarse-grained datasets UCF-101~\cite{soomro2012ucf101} and HMDB-51~\cite{6126543}, using Top-1 accuracy as metrics.
Specifically, FineGym includes hierarchical annotations at three semantic granularity: \textit{events}, \textit{sets}, and \textit{elements}.
At the finest level (elements), there are two versions of benchmarks, \textit{i.e.}, \textit{gym99} and \textit{gym288}, with 99 and 288 categories, respectively.
Note that all the experiments on FineGym are performed at the element level, but within different scopes.
FineDiving is a diving dataset comprising 3000 annotated clips with timestamps, encompassing 52 \textit{action} types, 29 \textit{sub-action} types, and 23 difficulty levels.

\vspace{-0.2em}
\subsubsection{Baselines.}
We employ the ViT~\cite{dosovitskiy2020image} extended model TimeSformer~\cite{bertasius2021space} as the backbone. 
The choice of hyperparameters remains as original.
We instantiate the SeFAR-S model based on ViT-Small, with the number of total parameters comparable to most previous Conv-based methods~\cite{han2020memory, xiong2021multiview, xu2022cross, xiao2022learning, 9904603, gowda2022learn2augment, dave2023timebalance}. 
Moreover, we implement the SeFAR-B model based on ViT-B, with more parameters. 
We configure the sampling combination by default as 
$\{2-2-4\}$ for SeFAR, as commonly used 8-frame input.

\vspace{-0.2em}
\subsubsection{Implementation Details.}
We employ our SeFAR using PyTorch, with input video frames resized and cropped to 224 × 224 pixels. 
We adopt the SGD optimizer and employ a learning rate of 0.005, momentum of 0.9, and weight decay of 0.001. 
The weights in Eq.\ref{eq4} are set as: $\lambda_{1} = \lambda_{2} = 2$.

\vspace{-0.2em}
\subsection{Main Results}
\label{sec:results}

The main quantitative results on the two fine-grained action recognition datasets, \textit{i.e.}, FineGym and FineDiving, are demonstrated in Tab.~\ref{tab:1}. 
We evaluate all the methods at different semantic granularities.
Specifically,
we first conduct experiments on Gym99 and Gym288.
Then, by narrowing the semantic scope, we focus on those element-level categories belonging to a specific event.
For instance, in Gym99, 25 classes belong to Uneven-bars (UB), while 35 classes are from Floor-exercise (FX).
Further, we delve into the finer granularity, collecting sampling within that same set in the same event. Here we get all the circles in UB-set1 (UB-S1) and all the jumps in FX-set1 (FX-S1) for evaluation.
We can observe that on both the FineGym and FineDiving, SeFAR-S significantly outperforms previous \textit{open-sourced} semi-supervised action recognition methods across all semantic granularities with moderate parameters. 
Additionally, when increasing the parameters comparative with SVFormer~\cite{xing2023svformer}, the larger model, SeFAR-B, performs even better. 
Both SeFAR-S and SeFAR-B display the effectiveness of our proposed SeFAR framework for addressing the challenging task.


Moreover, to further inspect the effectiveness and robustness of SeFAR, we conducted experiments on two classical coarse-grained action recognition datasets, UCF-101 and HMDB-51.
As shown in Tab.~\ref{tab:2},
SeFAR-B achieves approximately 3.3\% improvement on UCF101 and approximately 1.7\% improvement on HMDB51, achieving new state-of-the-art results compared with those competitive baselines. 

     

\vspace{-0.2em}
\subsection{Ablation Studies}
\label{sec:ablation}

To achieve an in-depth comprehension of our SeFAR framework, 
we perform ablation studies on the impact of each component, namely \textit{dual-level temporal elements modeling} (Dual-Ele), \textit{moderate temporal perturbation} (Mod-Perturb) and \textit{Adaptive Regulation} (Ada-Reg), as demonstrated in Tab.~\ref{tab:sub3.1}. Each module contributes significantly as an essential part of SeFAR.
Furthermore, we conduct a comprehensive analysis of the designs and choices of each proposed strategy or module. Details can be found as follows.

\vspace{-0.2em}
\subsubsection{Analysis of Dual-level Temporal Elements Modeling.} 
We first compare different combinations of sampled elements, each context element has varying temporal lengths, \textit{e.g.}, $4,6,8$. 
To facilitate comparison, we fix the length of the temporal fine-grained elements to be $2$, consistent with our default setting \{2-2-4\}.
Results are depicted in the left part of Fig.~\ref{fig:threshold}.
We can find that even with a limited input of only 6 frames, \textit{i.e.}, \{2-4\}, our proposed SeFAR surpasses the 8-frame input baseline SVFormer~\cite{xing2023svformer}. 
This observation justifies the capability of our \textit{dual-level temporal elements modeling} to capture abundant information details from video data, contributing to better discerning subtle differences among fine-grained actions. 
Additionally, it is noteworthy that increasing the number of the fine-grained elements, \textit{i.e.}, \{2-2-2-4\}, or extending the temporal length of the context element, \textit{i.e.}, \{2-2-6\} and \{2-2-8\}, all leads to performance improvements. This is attributed to the fact that more frames entail richer action information.

\begingroup
\vspace{-0.6em} 
\setlength{\tabcolsep}{2.6pt} 
\begin{table}
  \renewcommand{\arraystretch}{1}
  \centering
  \caption{\textbf{Ablations of different components with SeFAR}, where \ding{51} means ``w/". To adhere to the principle of consistency regularization in SSL, we employ strong augmentation consistent with SVFormer~\cite{xing2023svformer}, \textit{i.e.}, temporal warping, once our Mod-Perturb is eliminated.}
  \centering
  \vspace{-0.8em}
  \small
   \begin{tabular}{ccc|ccc}
     \hlineB{2.5}
     \textbf{Dual-Ele} & \textbf{Mod-Perturb} & \textbf{Ada-Reg} & \textbf{Gym99} & \textbf{Gym288} & \textbf{Diving}\\
     \hlineB{2}
     \rowcolor{yellow!10}
      \ding{55} & \ding{55} & \ding{55} & 32.6 & 22.7 & 60.4 \\
     \rowcolor{yellow!10}
       \ding{51}  & \ding{55} & \ding{55} & 34.8 & 25.4 & 64.6 \\
     \rowcolor{yellow!10}
       \ding{51} & \ding{51}  & \ding{55} & 35.9 & 26.6 & 67.4 \\
     \rowcolor{cyan!10}
      \ding{51}  &  \ding{51} & \ding{51}  & \textbf{36.7} & \textbf{27.8} & \textbf{72.2} \\
     \hlineB{2.5}
   \end{tabular}
  \label{tab:sub3.1}
  \vspace{-0.6em}
\end{table} 
\endgroup

\begingroup
\setlength{\tabcolsep}{1.6pt} 
\begin{table}[t]
  \renewcommand{\arraystretch}{1}
  \centering
  \caption{\textbf{Ablation of different temporal augmentations.} \textbf{S} and \textbf{O} denote the Speed- and Order-focused.}
  \centering
  \vspace{-0.8em}
  \small
   \begin{tabular}{l|c|ccccc}
     \hlineB{2.5}
     \textbf{Perturbation} & \textbf{S/O} & \textbf{Gym99} & \textbf{Gym288} & \textbf{Diving} & \textbf{G.-New} & \textbf{Sth.-Sth.}\\
     \hlineB{2}
     \rowcolor{yellow!10}
     Spatial-only &  & 34.2 & 24.4 & 67.9 & 45.6 & 39.4\\
     \hline
     \rowcolor{yellow!10}
     Slow (T-Drop) & S & 35.6 & 25.2 & 68.6 & 45.0 & 41.2 \\
     \hline
     \rowcolor{yellow!10}
       All shuffle & O  & 35.2 & 26.3 & 69.0 & 45.5 & 41.9\\
     \rowcolor{yellow!10}
       Local-shuffle & O  & 36.4 & 27.6 & 71.9 & 45.3 & 43.3\\
     \rowcolor{yellow!10}
        Warping & O  & 35.9 & 24.7 & 68.2 & 44.8 & 40.8\\
     \rowcolor{yellow!10}
       T-Half & O & 36.0 & 24.8 & 68.4 & 44.8 & 42.1\\
     \rowcolor{yellow!10}
       All reverse & O & 36.3 & 27.3 & 71.2 & 45.9 & 42.7\\
     \rowcolor{cyan!10}
      Mod-Perturb & O  & \textbf{36.7} & \textbf{27.8} & \textbf{72.2} & \textbf{46.2} & \textbf{44.9} \\
     \hlineB{2.5}
   \end{tabular}
  \label{tab:sub3.2}
  \vspace{-1.2em}
\end{table} 
\endgroup

\begingroup
\vspace{-0.6em} 
\setlength{\tabcolsep}{4pt} 
\begin{table}
  \renewcommand{\arraystretch}{1}
  \centering
  \caption{\textbf{Ablation of Pre-trained Visual Encoder.} We employ Vicuna-7B~\cite{chiang2023vicuna} as the base LLM, comparing SeFAR's features with the pre-trained features of commonly used visual encoders in MLLMs further fine-tuned on 5\% data (\textit{i.e.}, \begin{minipage}[b]{0.04\columnwidth}
		\raisebox{-.15\height}{\includegraphics[width=\linewidth]{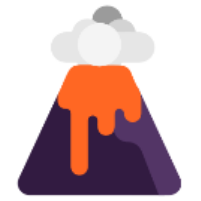}}
	\end{minipage}: LLaVA, \begin{minipage}[b]{0.05\columnwidth}
		\raisebox{-.25\height}{\includegraphics[width=\linewidth]{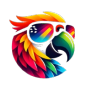}}
	\end{minipage}: VideoChat2, \begin{minipage}[b]{0.04\columnwidth}
		\raisebox{-.15\height}{\includegraphics[width=\linewidth]{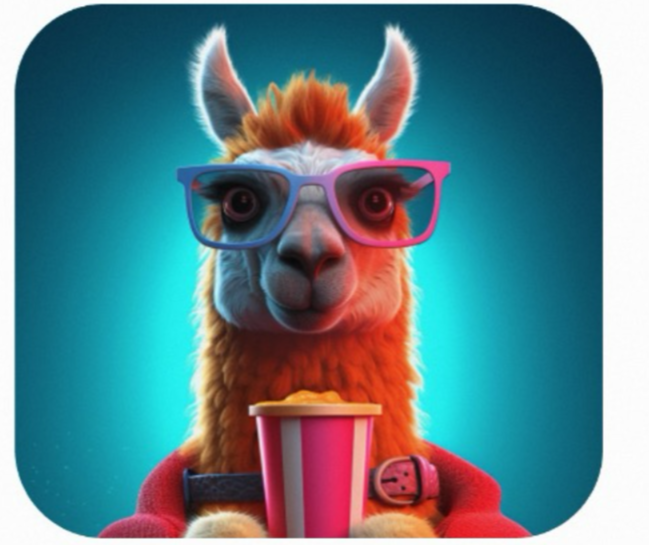}}
	\end{minipage}: VideoLLaMA, \begin{minipage}[b]{0.04\columnwidth}
		\raisebox{-.25\height}{\includegraphics[width=\linewidth]{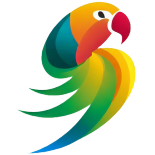}}
	\end{minipage}: VideoChat, and \begin{minipage}[b]{0.04\columnwidth}
		\raisebox{-.15\height}{\includegraphics[width=\linewidth]{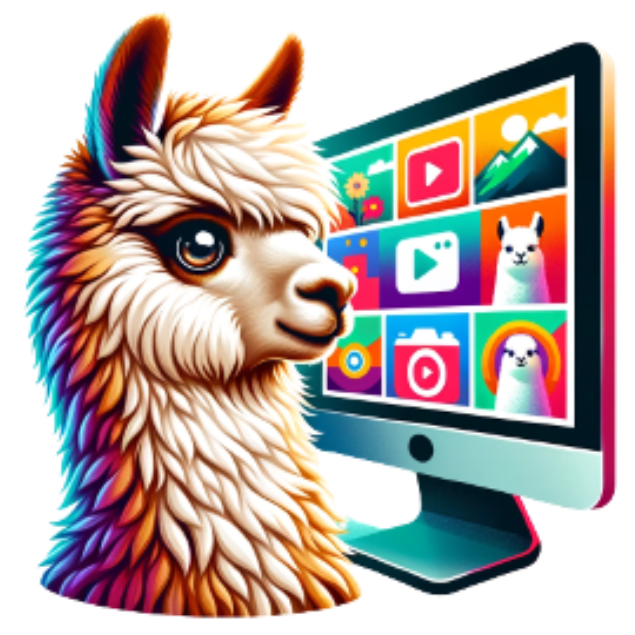}}
	\end{minipage}: VideoLLaVA)}
  \centering
  \vspace{-0.8em}
  \small
  
   \begin{tabular}{lc|cc}
     \hlineB{2.5}
     \textbf{Visual Encoder} & \textbf{MLLM} & \textbf{Gym-QA-99} & \textbf{Gym-QA-288} \\
     \hlineB{2}
     CLIP-ViT-L/16  & \begin{minipage}[b]{0.04\columnwidth}
		\raisebox{-.10\height}{\includegraphics[width=\linewidth]{LLaVA.png}}
	\end{minipage}, \begin{minipage}[b]{0.05\columnwidth}
		\raisebox{-.25\height}{\includegraphics[width=\linewidth]{VideoChat2.png}}
	\end{minipage} & 37.3 & 41.0 \\
     EVA-CLIP ViT-G/14 & \begin{minipage}[b]{0.04\columnwidth}
		\raisebox{-.15\height}{\includegraphics[width=\linewidth]{VideoLLaMA.png}}
	\end{minipage}, \begin{minipage}[b]{0.04\columnwidth}
		\raisebox{-.25\height}{\includegraphics[width=\linewidth]{VideoChat.png}}
	\end{minipage} & 43.7 & 44.8 \\
     ViT-L/14 & \begin{minipage}[b]{0.04\columnwidth}
		\raisebox{-.15\height}{\includegraphics[width=\linewidth]{VideoLLaVA.png}}
	\end{minipage} & 44.3 & 46.0 \\
      \hline
     \rowcolor{cyan!10}
      SeFAR (Ours) & - &\textbf{49.0} & \textbf{56.2} \\
     \hlineB{2.5}
   \end{tabular}
  \label{tab:mllm}
  \vspace{-1.8em}
\end{table}
\endgroup

\vspace{1.6em}
\subsubsection{Analysis of Moderate Temporal Perturbation.}
To better explore the impact of our proposed moderate temporal perturbation (Mod-Perturb), we first selected 40 classes of action pairs that are reversing to each other (\textit{e.g.}, ``giant circle backward" \textit{vs.} ``giant circle forward") from FineGym, forming a subset called \textbf{Gym-New} (G.-New).
As shown in Tab.~\ref{tab:sub3.2}, SeFAR also maintains superior performance even on such actions, as well as on the Something-Something V2 (Sth.-Sth.) dataset~\cite{goyal2017something}.
Furthermore, we compare our Mod-Perturb with other temporal perturbation strategies in both Speed- and Order-focused (\textit{e.g.}, slow-rate~\cite{singh2021semi}, temporal warping~\cite{xing2023svformer}, T-Drop and T-Half~\cite{zou2023learning}), the results can be found in Tab.~\ref{tab:sub3.2}.
We can observe that: 
1) Our Mod-Perturb exhibits superior stability and efficacy compared to other temporal augmentations and spatial-only (temporal information well-kept).
2) Spatial-only is less effective in Gym99 but outperforms most temporally augmented in Gym-New. This suggests that preserving accurate temporal information is crucial for more complex datasets, whereas reasonable temporal perturbations can enhance model stability in larger and more diverse datasets, and Mod-Perturb benefits from both.


\vspace{-0.2em}
\subsubsection{Analysis of Adaptive Regulation.} 
To justify the usefulness of our stabilizing coefficients for adaptive losses, we perform two analyses:
\ding{172} We compare this strategy with the fixed thresholding strategy widely used in the classical SSL method, the results are displayed in Fig.~\ref{fig:threshold} (b), showing our method is both stable and effective.
\ding{173} In Fig.~\ref{fig:threshold} (c), We demonstrate the unstable predictions provided by the teacher models for FAR.
Specifically, we randomly draw $30$ data samples from different datasets, UCF101, HMDB51, and FineGym, for the teacher model to offer predictions. 
The highly varying predictions on FineGym further justify the motivation of our stabilizing design for FAR.

\vspace{-0.2em}
\subsubsection{Analysis of SeFAR Features.} 
To further demonstrate the capability of SeFAR in enhancing MLLMs, we first constructed the \textbf{Gym-QA} dataset, which is derived from FineGym and presented in a multiple-choice format as illustrated in Fig.~\ref{fig:finegrained}. We then selected three widely used MLLM visual encoders, \textit{i.e.}, CLIP-ViT-L/16, EVA-CLIP ViT-G/14, and ViT-L/14). For fair comparisons, we conduct semi-supervised training on these backbones with 5\% labeling data from FineGym. Subsequently, we froze the weights of these encoders along with the weights from our 5\%-trained SeFAR, and fine-tuned the Q-former using 5\% of the annotated data from Gym-QA. As shown in Tab.~\ref{tab:mllm}, the SeFAR-empowered LLM significantly outperformed the other MLLM visual encoders on the Gym-QA task. This also mitigates the challenge of fine-tuning MLLMs in scenarios with low labeling rates.

\section{Conclusion}
In this work, we shed light on a more challenging and specific video understanding task, Semi-supervised Fine-grained Action Recognition (FAR).
To tackle the unique challenges that emerged,
we propose a new framework, SeFAR, which adopts ideas from the FixMatch setting and possesses innovative components delicately devised for FAR.
Specifically, SeFAR is distinguished due to the following designs:
1) \textit{Dual-level temporal elements modeling} is used to mine visual cues more thoroughly and capture rich temporal dynamics better;
2) \textit{Augmentation via moderate temporal perturbation} is to produce temporally strong-distorted samples for weak-to-strong consistency regularization;
3) \textit{Stabilizing Optimization via Adaptive Regulation} is to address the issue of large uncertainty in model predictions. To highlight, SeFAR also demonstrates superior performance in empowering MLLM's fine-grained visual understanding capability.
SeFAR not only outperforms all the baselines largely on two representative FAR datasets, FineGym and FineDiving,
but also achieve new state-of-the-art results on classical benchmarks (\textit{i.e.}, UCF101 and HMDB51).
 Comprehensive analysis and Extensive ablation studies further justify the effectiveness of our framework design.

\section{Acknowledgments}
This work was founded by the National Natural Science Foundation of China (NSFC) under Grant 62306239, and was also supported by National Key Lab of Unmanned Aerial Vehicle Technology under Grant WR202413.

\bibliography{aaai25}


\clearpage
\appendix
\newpage

\section{Appendix}
\subsection{Introduction}
The content of our Appendix is organized as follows:

\noindent\ding{224} In Section A, we present the data processing employed in our SeFAR framework, as well as baseline analysis;

\noindent\ding{224} In Section B, we expound upon the categorical analysis of model uncertainty;


\noindent\ding{224} In Section C, we provide more discussions regarding our SeFAR; 

\noindent\ding{224} In Section D, we present detailed information regarding the newly built Gym-QA and Gym-New datasets.


\subsection{A. Data Processing and Baseline Analysis}
\subsubsection{Data Processing.}In order to ensure a rigorous and equitable comparison, we adopt identical data processing procedures and input formats across both SeFAR and the baseline methods.
It is noteworthy that the input data format utilized in our experiments may not be the same as the original versions presented in FineGym~\cite{shao2020finegym} and FineDiving~\cite{xu2022finediving} since we conduct experiments at the finest level within these two datasets.
We release the data pre-processing scripts together with the whole project code for the convenience of future work.

\subsubsection{Baseline Analysis.} Semi-supervised fine-grained action recognition is a challenging task that has not been previously explored. This is evident from the experimental results (\textit{e.g.}, Tab.~\ref{tab:1}), which show that models designed for coarse-grained action perform poorly on fine-grained actions. It is important to clarify that the baselines compared in Tab.~\ref{tab:1} were evaluated and tested by us on the FineGym and FineDiving datasets for the first time, rather than by previous studies. The reason for including only three baselines in this comparison is that, although many studies have explored semi-supervised coarse-grained action recognition (\textit{e.g.}, baselines in Tab.~\ref{tab:2}), only the three works presented in Tab.~\ref{tab:1} are open-source and reproducible. We also attempted to reproduce models from non-open-source works based on their methodology sections, but unfortunately, the results we obtained differed from those reported in their original papers. 

Notably, we have identified an impressive concurrent work, FinePseudo~\cite{dave2025finepseudo}, which is dedicated to addressing the problem of semi-supervised fine-grained action recognition. We will give it further attention and exploration in our future work.

\subsection{B. Visualization of Model Uncertainty}
As highlighted by~\cite{rizve2021defense}, the escalation of uncertainty within model predictions inversely impacts the model's reliability. A parallel phenomenon is discernible in our exploration of semi-supervised fine-grained action recognition. Specifically, we conducted a random sampling of 1000 data points from various datasets and employed the Teacher model to predict each set of 1000 data points, subsequently evaluating the accuracy of the Teacher model. The left panel of Fig.~\ref{fig:prediction} illustrates the correlation between the confidence level associated with the Teacher model's predictions and their corresponding accuracy; notably, predictions characterized by heightened confidence demonstrate augmented accuracy. Conversely, the right panel of Fig.~\ref{fig:prediction} depicts the connection between the standard deviation of predictions generated by the Teacher model and their accuracy; a diminished variance in predictions is concomitant with heightened accuracy. This visual representation corroborates the rationale underpinning our conceptualization of the \textit{Adaptive Regulation}.

\begin{figure}
    \centering
    \small
   \includegraphics[width=1\linewidth]{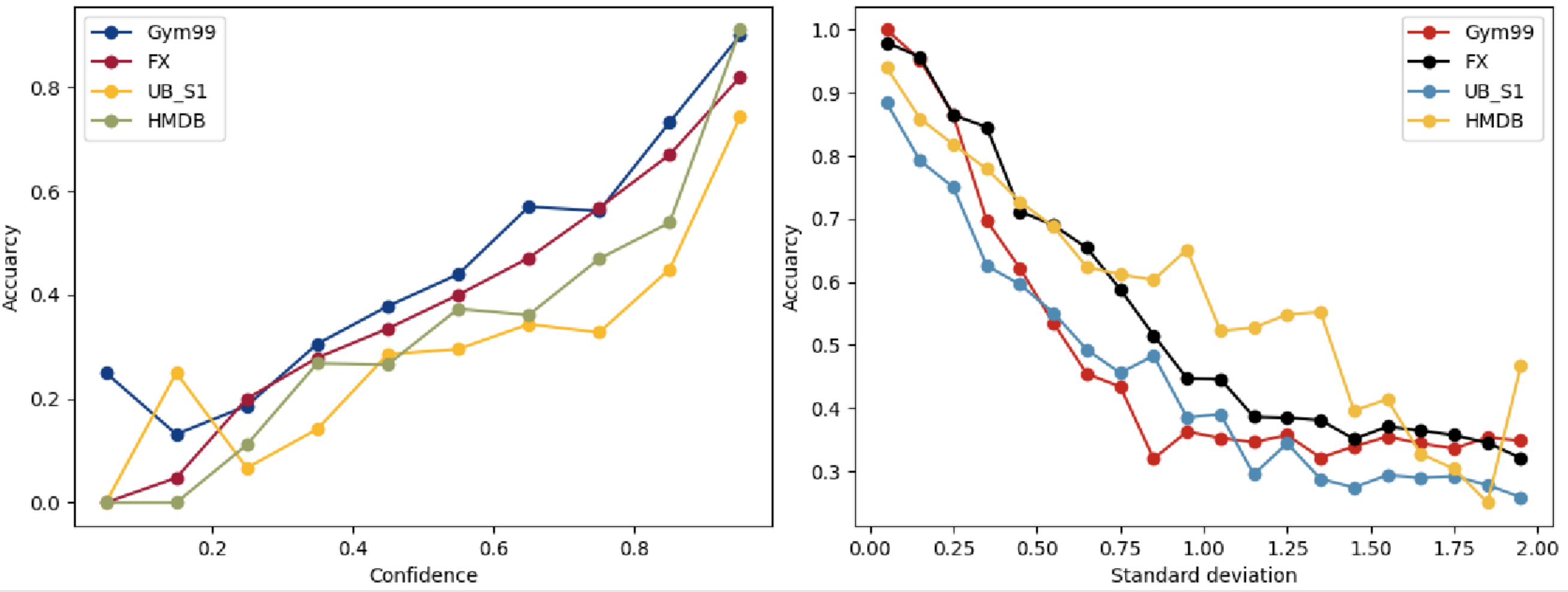}
   \vspace{-1.9em}
    \captionof{figure}{\label{fig:prediction}
    The relationship between the Teacher model's prediction accuracy and its confidence (left), as well as its standard deviation (right).
    }
\end{figure}


\subsection{C. More Discussions on SeFAR}

In this section, we further discuss the following research questions ($\mathcal{RQ}$):
\begin{enumerate}[start=1,label={\bfseries $\mathcal{RQ}$\arabic*:},leftmargin=3em,itemsep=-1mm]
\item How does each module contribute to enhancing fine-grained action recognition?
\item How does the \textit{dual-level temporal elements modeling} differ from previous modeling strategies?
\item Will the \textit{moderate temporal perturbation} alter the actions?
\item Can the \textit{adaptive regulation} be effective under more challenging conditions?
\item Does the teacher model’s prediction frequency affect performance?
\item What are the limitations of the current approach, and what directions should future research take?
\end{enumerate}

\subsubsection{Module Analysis of SeFAR ($\mathcal{RQ}$1):}
\ding{182} \textit{Dual-level temporal elements modeling}: As discussed earlier, fine-grained actions, compared to coarse-grained actions, not only rely heavily on global semantic context but also contain richer visual detail, presenting unique challenges. The dual-level temporal elements we designed divide a video into two hierarchical levels (\textit{i.e.}, fine-grained elements $p_i$, and context element $p^{\textbf{context}}$). This provides multi-granular information for fine-grained action recognition, allowing the model to capture features at different temporal scales (\textit{e.g.}, varying numbers of giant swings), and offering diverse representations for actions of different durations.
\ding{183} \textit{Moderate temporal perturbation}: In semi-supervised learning, data augmentation is essential for consistency regularization, which leads to more stable and superior model performance. Traditional coarse-grained action recognition often uses spatial augmentations that may disrupt critical details needed for fine-grained actions. For example, while coarse-grained actions like ``running" can be recognized even with masked frames, fine-grained actions are characterized by complexity and coherence. Therefore, we focus on temporal augmentations in this work. As shown in Tab.~\ref{tab:sub3.2}, excessive perturbations can disrupt sequence information, hindering the model's ability to capture subtle action differences. Our experiments show that sequence reversal provides strong perturbations while preserving action continuity, making them more effective for temporal augmentation. Additionally, our moderate temporal perturbation retains global context, enabling the model to benefit from augmentation while maintaining a coherent understanding of actions.
\ding{184} \textit{Adaptive regulation}: In fine-grained action recognition, subtle differences between similar actions (\textit{e.g.}, examples in Fig.~\ref{fig:finegrained}) can lead to significant fluctuations in the predictions made by the Teacher model, particularly in a semi-supervised setting, as illustrated in Fig.~\ref{fig:threshold}. The adaptive regulation strategy plays a crucial role in stabilizing the training process by automatically adjusting the weights of the loss functions based on the distribution of the Teacher model's predictions, which is essential for effective semi-supervised fine-grained action recognition.

\begingroup
\setlength{\tabcolsep}{3pt} 
\begin{table}[t]
  \renewcommand{\arraystretch}{1.2}
  \centering
  \caption{\textbf{Ablation of different labeling rates.} The first two raw demonstrate our SeFAR w/o and w/ the Adaptive Regulation (Ada-Reg) respectively. The third raw further shows the performance increase rates at different labeling rates.}
  \centering
  \vspace{-0.8em}
  \small
   \begin{tabular}{l|ccccc}
     \hlineB{2.5}
     \multirow{2}{*}{\textbf{Method}} & \multicolumn{5}{c}{\textbf{FineDiving}}\\
     \cline{2-6}
      & \textbf{1\%} & \textbf{3\%} & \textbf{5\%} & \textbf{7\%}& \textbf{10\%}\\
     \hlineB{2}
     \rowcolor{yellow!10}
       SeFAR w/o Ada-Reg & 61.5 & 64.6 & 67.2 & 69.7 & 73.4 \\
     \rowcolor{cyan!10}
      SeFAR & \textbf{66.3} & \textbf{69.5} & \textbf{72.2} & \textbf{74.6} & \textbf{78.4} \\
      Increase (\%) & 7.8\%$\uparrow$ & 7.6\%$\uparrow$ & 7.4\%$\uparrow$ & 7.0\%$\uparrow$ & 6.8\%$\uparrow$ \\
     \hlineB{2.5}
   \end{tabular}
  \label{tab:suppl_regulation}
  \vspace{-0.4em}
\end{table} 
\endgroup

\begingroup
\setlength{\tabcolsep}{4pt} 
\begin{table}[t]
  \renewcommand{\arraystretch}{1.2}
  \centering
  \caption{Deeper comparison of temporal augmentations.}
  \centering
  \vspace{-0.8em}
  \small
   \begin{tabular}{l|c|cccc}
     \hlineB{2.5}
     \textbf{Perturbation} & \textbf{Speed/Order} & \textbf{FX} & \textbf{10m} & \textbf{UB-S1} & \textbf{5253B} \\
     \hlineB{2}
     \rowcolor{yellow!10}
     Slow-rate & Speed & 22.4 & 81.2 & 35.6 & 92.8 \\
     \rowcolor{yellow!10}
     T-Drop & Speed & 22.4 & 81.2 & 35.6 & 92.8\\
     \hline
     \rowcolor{yellow!10}
       All shuffle & Order  & 23.5 & 82.8 & 36.1 & 93.5\\
     \rowcolor{yellow!10}
       Local-shuffle & Order  & 23.0 & 84.1 & 36.5 & 94.9 \\
     \rowcolor{yellow!10}
        Warping & Order  & 23.4 & 81.9 & 34.7 & 92.9 \\
     \rowcolor{yellow!10}
       T-Half & Order & 23.3 & 83.0 & 35.3 & 93.4 \\
     \rowcolor{yellow!10}
       All reverse & Order & 23.6 & 83.7 & 35.5 & 95.1 \\
     \rowcolor{cyan!10}
      Mod-Perturb & Order  & \textbf{23.8} & \textbf{85.5} & \textbf{36.6} & \textbf{96.4}\\
     \hlineB{2.5}
   \end{tabular}
  \label{tab:app_aug}
  \vspace{-0.4em}
\end{table} 
\endgroup

\subsubsection{Dual-level Temporal Elements Modeling ($\mathcal{RQ}$2):}
Sampling at different temporal scales is actually not a new approach in action recognition. However, unlike previous methods, \textit{e.g.}, TPN~\cite{yang2020temporal}), which model at the \textit{feature} level and sample once at each level, following an ``$L^{1} < L^{2}... <L^{N}$" hierarchy for an N-level pyramid, often leading to high frame sampling and computational demands, our dual-level temporal elements modeling represents different video speeds through multi-level sampling at the input stage. We employ multiple fine-grained elements, each with the same number of frames (\textit{i.e.}, 2), and a single context element to capture local and global features, respectively. This design allows us to achieve better performance while minimizing the total number of sampled frames.

To achieve a deeper comparison with other temporal perturbations, we assessed each method on sub-tasks involving the recognition of elements with an event (FX, 10m) and within a set (UB-S1, 5253B) using 10\% labeled data as shown in Tab.~\ref{tab:app_aug}. Consistent with the results in the main text (Tab.~\ref{tab:sub3.2}), our proposed moderate temporal perturbation (Mod-Perturb) consistently outperformed all other strategies across all sub-tasks, demonstrating its superior efficacy.


\begin{figure}[t]
    \centering
    \small
   \includegraphics[width=1\linewidth]{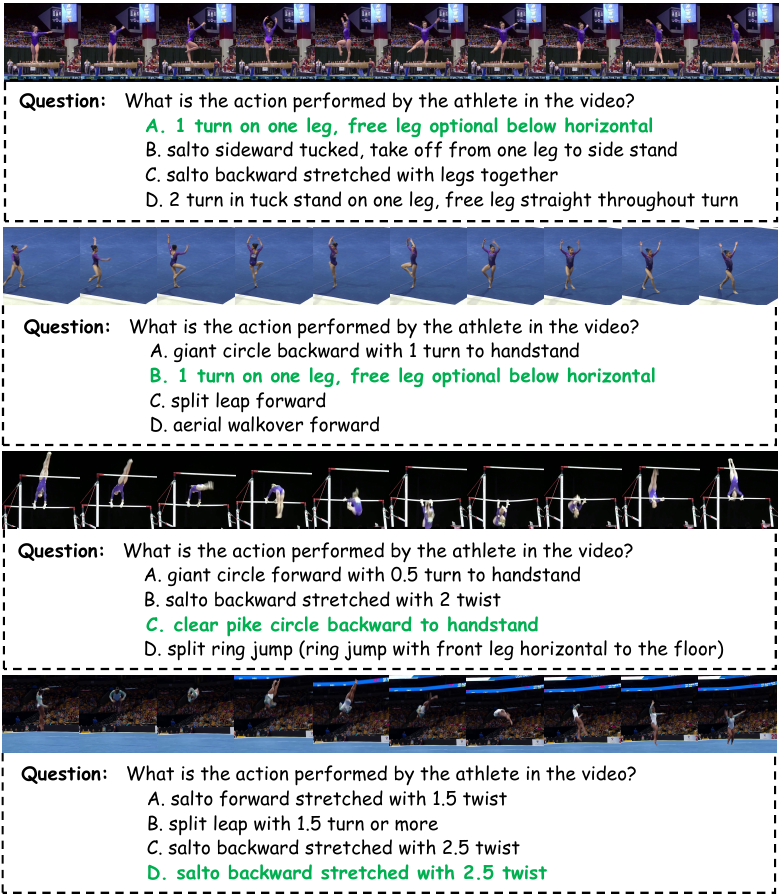}
   \vspace{-1.9em}
    \captionof{figure}{\label{fig:gym_qa}
    Examples of Gym-QA
    }
    \vspace{-1em}
\end{figure}

\begin{figure}[t]
    \centering
    \small
   \includegraphics[width=1\linewidth]{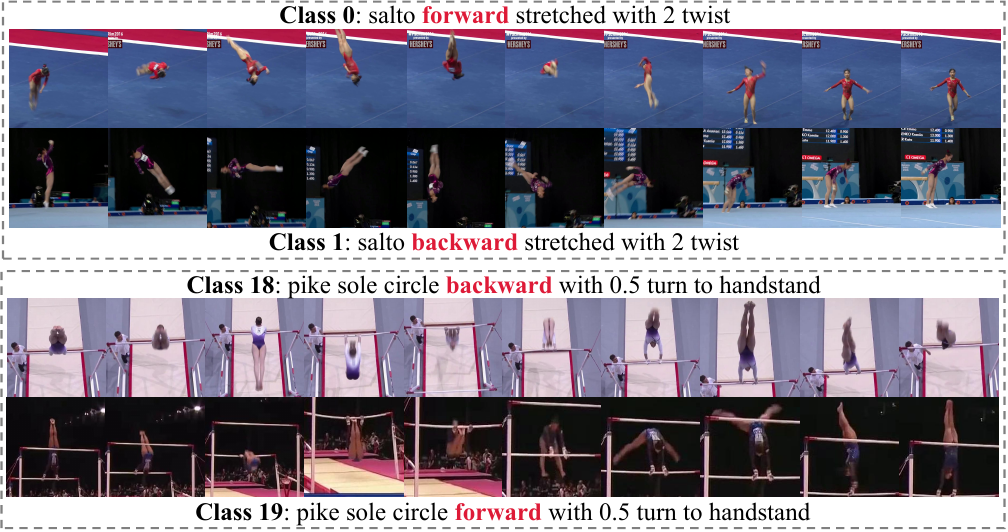}
   \vspace{-1.9em}
    \captionof{figure}{\label{fig:gym_new}
    Examples of Gym-New
    }
    \vspace{-1em}
\end{figure}

\begin{figure*}[t]
    \centering
    \small
   \includegraphics[width=1\linewidth]{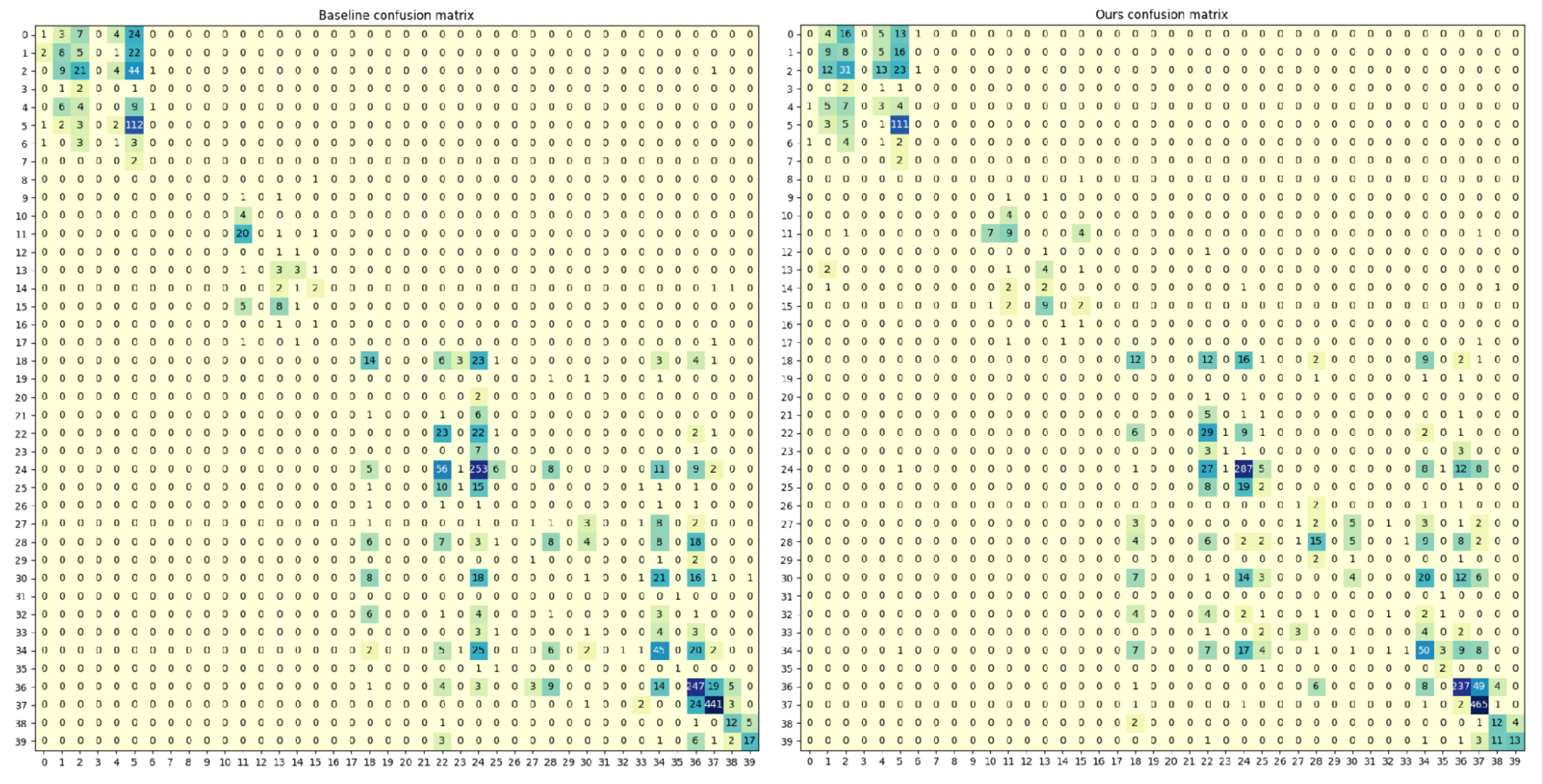}
   \vspace{-1.9em}
    \captionof{figure}{\label{fig:confusion}
    Confusion matrix of baseline (left) and ours (right) on Gym-New 10\%, where the horizontal coordinate represents the predicted label and the vertical coordinate represents the true label. The labels corresponding to actions are shown in Fig.~\ref{fig:cate}.
    }
    \vspace{-1em}
\end{figure*}

\subsubsection{Potential Action Directionality Changes ($\mathcal{RQ}$3):}
Human actions are inherently complex to a certain extent~\cite{shao2018find, shao2020intra}. Intuitively, the reversal of action videos introduces challenges related to the directionality of actions (\textit{e.g.}, ``sitting down" \textit{vs.} ``standing up"). We have taken this into account in our dual-level temporal elements modeling design, which includes both fine-grained elements containing local details and context elements capturing global information. During temporal perturbation, we only reverse the fine-grained elements, preserving the original temporal order in the context elements. This allows us to achieve consistency regularization through temporal perturbation while maintaining the original global temporal structure, which differs significantly from complete reversal and previous temporal augmentation methods applied at the video level. This also indicates that our dual-level temporal elements modeling is coupled with moderate temporal perturbation, rather than being a simple modular combination. Furthermore, as demonstrated in Tab.~\ref{tab:sub3.2}, we validated this approach by constructing a variant of the FineGym dataset composed of completely \textit{opposite action pairs}, named Gym-New, for experimentation. The results further confirm that for fine-grained action recognition tasks, which require temporal and spatial coherence, common temporal augmentation strategies may disrupt this coherence, whereas our moderate temporal perturbation maintains coherence while introducing significant temporal disturbance.

\begin{figure*}[t]
 \vspace{-1em}
    \centering
    \small
   \includegraphics[width=1\linewidth]{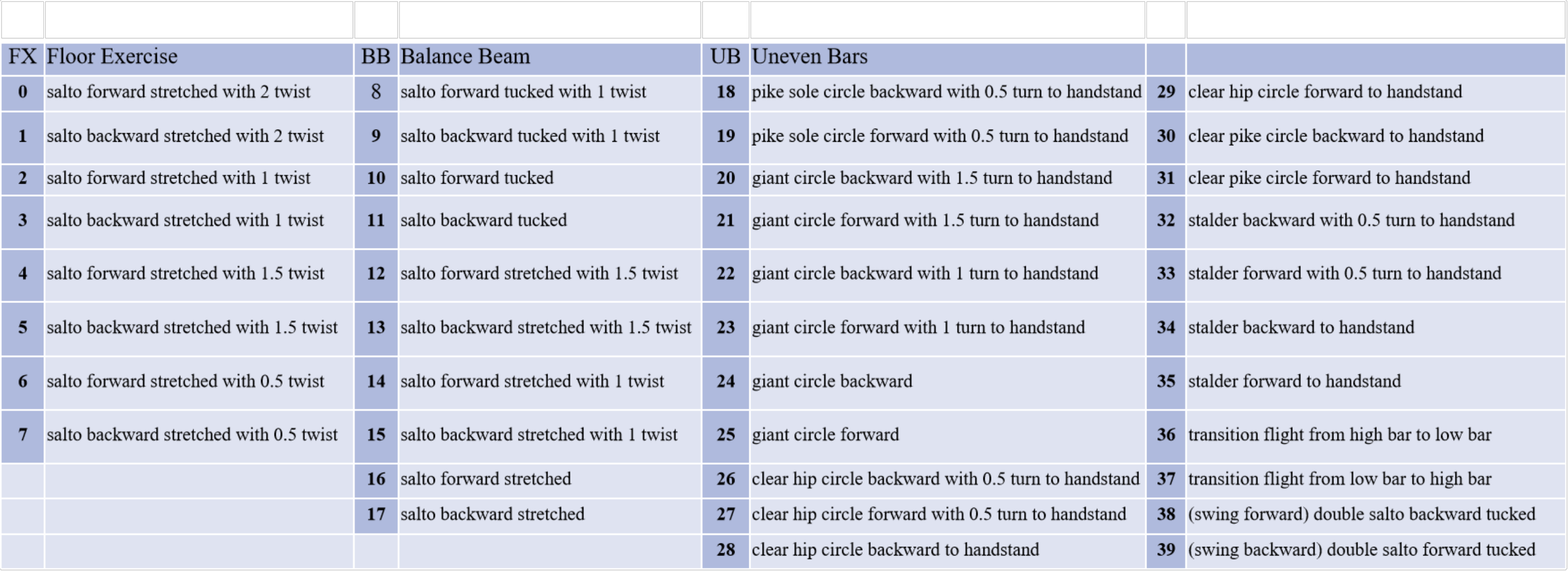}
   \vspace{-1.9em}
    \captionof{figure}{\label{fig:cate}
    Labels corresponding to actions in Gym-New.
    }
\end{figure*}

\subsubsection{Adaptive Regulation ($\mathcal{RQ}$4):}
With the continuous advancement of deep learning~\cite{chen2024finecliper, yan2024urbanclip, chen2024gaussianvton, ma2024beyond, huang2024crest, yan2024errorradar, shu2024cmt, wu2024fsc}, the data-hungry paradigm of fully supervised learning has increasingly revealed certain limitations. Unlike the extensively studied fully-supervised setting, semi-supervised learning typically operates with a label rate ranging from 1\% to 10\%, making it particularly suitable for tasks like fine-grained action recognition that require high-quality data. However, low label rates can lead to instability during training, as discussed in the main text. To address this challenge, we designed the Adaptive Regulation process. In a semi-supervised setting, lower label rates present greater difficulties. Therefore, to further explore the potential of our strategy, we conducted experiments under varying label rates, as shown in Tab.~\ref{tab:suppl_regulation}. The results demonstrate that as the label rate decreases, the performance enhancement provided by adaptive regulation becomes more pronounced, further validating that our strategy can effectively maintain strong performance under more challenging conditions.

\begingroup
\setlength{\tabcolsep}{2pt} 
\begin{table}[t]
  \renewcommand{\arraystretch}{1.2}
  \centering
  \caption{Computation analysis of teacher model predictions. Time shown in (ms).}
  \centering
  \vspace{-0.8em}
  \small
   \begin{tabular}{l|cccccc}
     \hlineB{2.5}
     \textbf{Prediction Times} & \textbf{1} & \textbf{2} & \textbf{5} & \textbf{10} & \textbf{15}  & \textbf{20}\\
     \hlineB{2}
     \rowcolor{yellow!10}
     Teacher time / Iter. & 29.9 & 68.5 & 75.8 & 160.4 & 260.1 & 361.3 \\
     \rowcolor{yellow!10}
     Total time / Iter.& 982.8 & 991.6 & 1005.1 & 1080.7 & 1220.6 & 1417.6 \\
     Portion (\%) & 3.0 & 6.9 & 7.5 & 14.8 & 21.3 & 25.5\\
     \hline
     \rowcolor{yellow!10}
     Accuracy (\%) & - & 35.3 & 36.2 & 36.7 & 36.8 & 37.0\\
     \hlineB{2.5}
   \end{tabular}
  \label{tab:app_eff}
  \vspace{-0.4em}
\end{table} 
\endgroup

\subsubsection{Efficiency of Teacher Model Prediction ($\mathcal{RQ}$5):}
During inference, only the student model is used, incurring no additional computational cost from the teacher model. In the training phase, as shown in Tab.~\ref{tab:app_eff}, we conduct further analysis focusing on \ding{182} \textit{Time Cost}: Teacher prediction time increases with more predicted but remains a small fraction of total training time (\textit{e.g.}, 14.8\% at $10$ predictions). This efficiency is achieved as teacher predictions are parallelized and do not involve gradient computations.
\ding{183} \textit{Accuracy Impact}: Model accuracy improves with the number of predictions, tending to saturate around $10$ predictions. Therefore, we set the number of teacher predictions to $10$ to balance performance and computational efficiency.

\subsubsection{Limitation and Future Work ($\mathcal{RQ}$6):}
In this work, we introduce SeFAR to address the challenging task of semi-supervised fine-grained action recognition for the first time, achieving superior performance with the aid of our carefully designed modules. This advancement establishes a robust baseline for future research. However, one limitation of this study is that we focused on temporal augmentation to emphasize its importance in fine-grained action understanding, while neglecting further exploration of spatial augmentation. We plan to address this in future work.

Another potential limitation is that our core modules rely solely on RGB video input, overlooking the contribution of multimodal information in visual tasks. While we acknowledge that multimodal inputs, \textit{e.g.}, pose and textual descriptions, can significantly enhance model performance, we think that for the specific task of fine-grained action recognition—where data collection and annotation are particularly challenging—relying on such inputs could limit the model’s generalizability. Moreover, the extraction and annotation of fine-grained action-related pose and textual descriptions pose significant challenges due to their complex nature and the domain-specific knowledge required.

With the advancement of generative models~\cite{chen2024omnicreator, zheng2024videogen}, we will strive to overcome these limitations in future work and further explore advanced models' fine-grained visual understanding and generation capabilities.

\subsection{D. Gym-QA and Gym-New}

\subsubsection{Gym-QA.} To facilitate the evaluation of MLLMs in fine-grained action understanding, we adapted the FineGym dataset into a multiple-choice format, creating the Gym-QA dataset, as illustrated in Fig.~\ref{fig:gym_qa}. Following the coarse-grained action recognition paradigm from VideoChat2~\cite{li2024mvbench}, we posed the question: ``What action is the athlete performing in the video?" The answer options included one correct label and three distractor labels from the FineGym dataset.

\subsubsection{Gym-New.} As demonstrated in Fig.~\ref{fig:gym_new}, the Gym-New dataset is created by selecting direction-opposite action pairs from FineGym. This aims to provide a more challenging environment for fine-grained action understanding, further testing the temporal perturbation that is the focus of our work.

To delve deeper into the temporal directionality of actions, as illustrated in Fig.~\ref{fig:confusion}, we present the confusion matrices of our baseline, namely SVFormer~\cite{xing2023svformer} (Left), and our proposed SeFAR (Right) applied to FineGym-New dataset. Our method capitalizes on \textit{dual-level temporal elements modeling}, which yields diverse temporal features, and \textit{moderate temporal perturbation}, which enhances the model's focus on temporal feature modeling. This leads to two notable improvements over the baseline: \textbf{a)} Our method effectively mitigates the impact of class imbalance, manifesting in a significant increase in the accuracy of under-represented classes; \textbf{b)} Our approach minimizes confusion between actions with opposing temporal directions (\textit{e.g., ``forward" vs. ``backward"}), while also reducing confusion among similar actions, \textit{e.g., ``giant circle backward with 1 turn to handstand" vs. ``giant circle backward"}.

\end{document}